\documentclass[runningheads]{llncs}

\usepackage{eccv}

\usepackage{eccvabbrv}

\usepackage{graphicx}
\usepackage{booktabs}

\usepackage[accsupp]{axessibility}  % Improves PDF readability for those with disabilities.

\usepackage{hyperref}

\usepackage{orcidlink}
\usepackage[capitalize]{cleveref}
\usepackage{multirow}
\Crefname{figure}{Fig.}{Figs.}
\Crefname{section}{Sec.}{Secs.}
\Crefname{equation}{Eq.}{Eqs.}
\newcommand{\Bezier}{B\'{e}zier\xspace}
\newcommand{\ourmethod}{TopoGPT\xspace}

\begin{document}

\title{Generative Lane Topology Reasoning via Autoregressive Model with Geometry Prior}

\titlerunning{Generative Lane Topology Reasoning with Geometry Prior}

\author{Jiahui Fu\inst{1}\orcidlink{0000-0002-7475-6770} \and
Zehao Huang\orcidlink{0000-0003-1653-208X} \and Han Li\inst{1}\orcidlink{0000-0001-9368-826X} \and
Naiyan Wang\orcidlink{0000-0002-0526-3331} \and Si Liu\inst{1}\orcidlink{0000-0002-9180-2935}}

\authorrunning{J.~Fu et al.}

\institute{Institute of Artificial Intelligence, Beihang University\\
\email{\{jiahuifu, lihan0620, liusi\}@buaa.edu.cn}\\
\email{\{zehaohuang18, winsty\}@gmail.com}
}

\maketitle

\begin{center}
    \centering
    \includegraphics[width=0.8\textwidth]{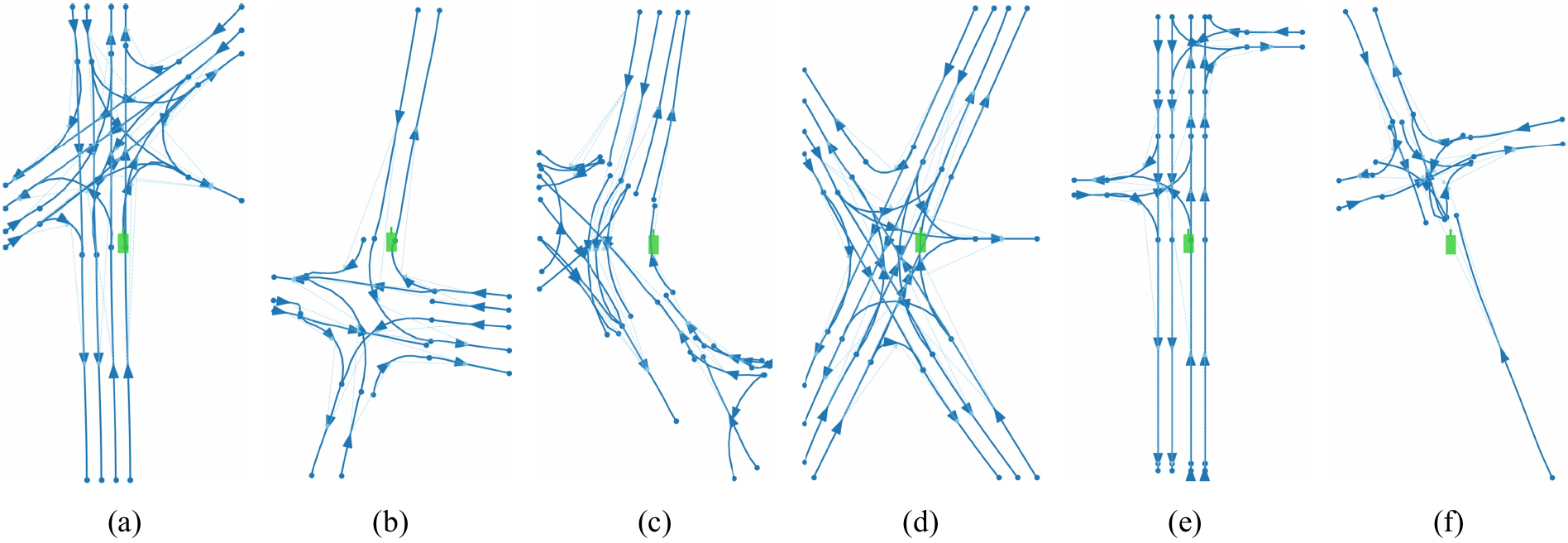}
    \captionsetup{
        type=figure,
        width=0.8\textwidth,
    }
    \captionof{figure}{\textbf{Samples of predicted lane graphs.} Given predicted lane graphs, human intuition can readily distinguish realistic structures from unrealistic ones without visual input. Test your intuition.\protect\footnotemark}
    \label{fig:teaser}
\end{center}
\footnotetext{\scriptsize (a,d,e) from our predictions w\!/ geometry prior; (b,c,f) from methods w\!/o geometry prior.}

\begin{abstract}
Lane topology reasoning aims to construct a lane graph from onboard sensor observations. Existing methods follow a detection and association paradigm that treats each lane instance independently, leading to geometric inconsistency at connected endpoints and incomplete graphs due to visual occlusions. 
To address these issues, we propose TopoGPT, a generative framework that learns the geometry prior from typical lane graph structures through autoregressive sequence modeling. 
Specifically, we construct a large-scale map dataset comprising $3.3$M scenes. For each lane graph, a lane tokenizer serializes it into discrete tokens, while a scene context encoder converts it into a rasterized image and extracts global features as scene tokens. We pre-train an autoregressive lane sequence transformer via scene-conditioned next-token prediction, endowing the model with the geometry prior over lane graph structures.
Building upon this prior, a perception adapter aligns BEV features from multi-view images with the pre-trained scene condition, transferring the learned geometry prior to sensor-based lane graph prediction.
On the OpenLane-V2 benchmark, TopoGPT outperforms existing methods by an average of $+6.4$ on lane-level and $+11.6$ on point-level metrics, and produces geometrically consistent and structurally complete lane graphs. Our project page is available at \url{https://buaa-colalab.github.io/topogpt_page}.
\end{abstract}

\section{Introduction}
\label{sec:intro}
Lane topology reasoning~\cite{Openlanev2, LaneGraphNet} constructs a vectorized lane graph from onboard sensor observations by jointly detecting centerlines and inferring their topological relations (\ie, predecessor/successor). This converts raw perception into a structured, navigable representation for autonomous driving.

\begin{figure}[t]
  \centering
  \includegraphics[width=1.0\linewidth]{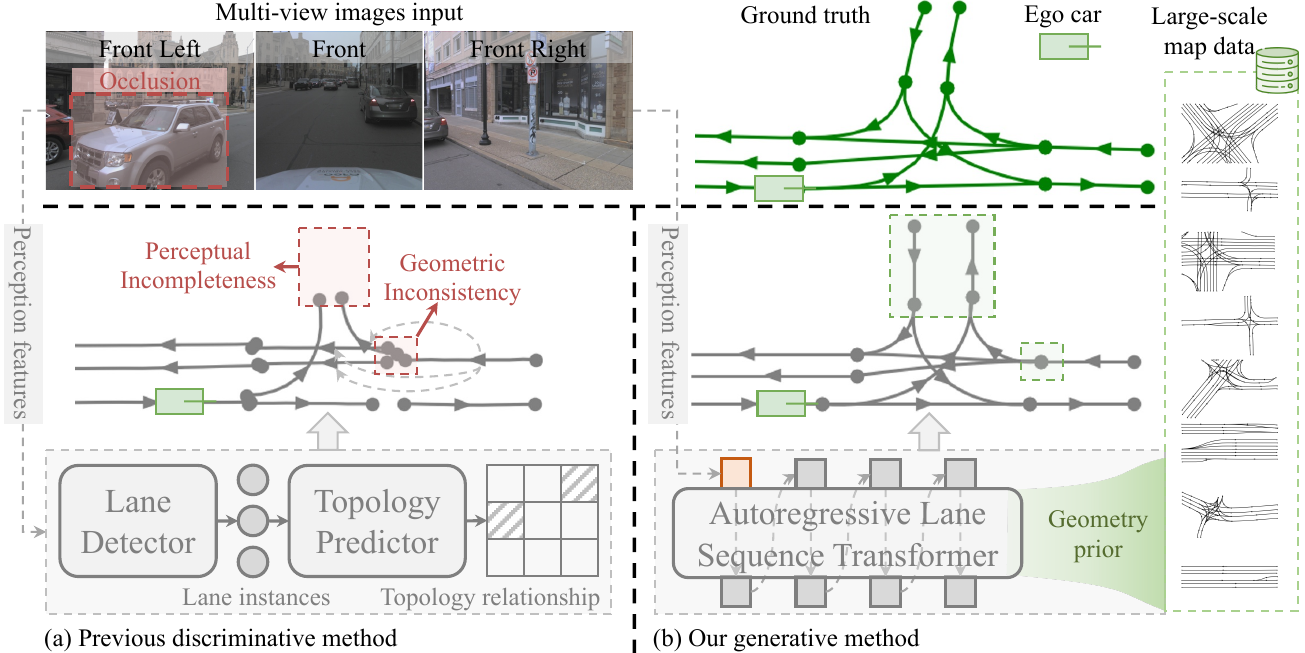}
  \caption{Given multi-view images: (a) Previous discriminative methods suffer from endpoints misalignment and missed lanes under occlusion. (b) Our generative model learns the geometry prior from large-scale map data and predicts contiguous and complete lane graphs conditioned on image-derived perception tokens.}
  \label{fig:motivation}
\end{figure}

However, existing methods~\cite{TopoNet, TopoMLP, TopoLogic, TopoPoint} typically decouple lane detection from topology prediction in a first-detect-then-associate paradigm. As shown in~\cref{fig:motivation} (a), a lane detector first predicts lane instances, after which a topology predictor estimates dense pairwise relationships. This two-stage discriminative design inevitably introduces two issues.
(1)~\textit{Geometric Inconsistency}: Endpoints of topologically connected centerlines are often misaligned, as the decoupled topology stage cannot retroactively refine centerline geometry.
(2)~\textit{Perceptual Incompleteness}: The lane graph is bounded by detector recall, which degrades under occlusion and limited sensing range. The subsequent association stage cannot recover missed instances, yielding a fragmented graph.

In contrast to generic object detection, where instances are often spatially independent, centerlines in a lane graph are governed by strict topological constraints inherent to road layout. 
As illustrated in~\cref{fig:teaser}, humans can readily judge whether a lane graph is structurally realistic even without surrounding camera images, owing to intrinsic geometry priors about common road structures. 
This motivates us to investigate whether a model can be endowed with a similar capability, enabling predictions that conform to such a prior distribution.
To this end, we introduce the \textbf{G}enerative \textbf{P}re-trained \textbf{T}ransformer for Lane \textbf{Topo}logy~(\textbf{\ourmethod}), which models lane graphs as a global structure rather than independent instances. 
Leveraging millions of open-source map scenes, we pre-train \ourmethod to learn the geometry prior and capture global structures of lane graphs. 
Because map-only data does not require accurately synchronized images and map annotations, it can be collected from diverse HD map sources at a much larger scale than sensor-map paired data.
Building upon this learned prior, we fine-tune the model to produce geometrically consistent and structurally complete predictions conditioned on online sensor observations.

As illustrated in~\cref{fig:motivation}~(b), we design an Autoregressive Lane Sequence Transformer~(ALST) to learn the geometry prior over lane graphs via generative sequence modeling. To serialize a lane graph, a {\Bezier}-based lane tokenizer converts each centerline into a group of discrete lane tokens. These groups are then concatenated according to a predefined spatial order, forming a complete lane token sequence that can directly reconstruct the original lane graph. Furthermore, we rasterize each lane graph into a pseudo-image and use a scene context encoder to extract global features as scene tokens. To support large-scale pre-training, we develop a dedicated data-processing pipeline that standardizes lane graphs from diverse data sources into a unified representation, resulting in a dataset of \(3.3\)M scenes, on which we pre-train ALST via next lane token prediction conditioned on scene tokens.
After pre-training, we fine-tune it on sensor-map paired data. In particular, a Bird's-Eye-View (BEV) encoder extracts BEV tokens from multi-view images, and a perception adapter projects them into the same condition token space as the scene tokens. Meanwhile, we apply LoRA-based fine-tuning to adapt ALST to BEV token conditioning.
Benefiting from the geometry prior learned during pre-training, our method produces geometrically continuous connections between lane instances, allowing accurate topology prediction via a simple distance-based endpoint association criterion. Moreover, even under occlusion, since each lane token is generated conditioned on both perception tokens and preceding lane tokens, the model can infer reasonable topology in occluded regions to complete missing lanes consistent with the lane graph structure.
We benchmark TopoGPT on the OpenLane-V2~\cite{Openlanev2} dataset against state-of-the-art discriminative methods. On subset A, TopoGPT consistently outperforms prior methods in spatial localization accuracy (reflected by $+6.4$ mean gain in the lane-level metric) and geometric structure realism (reflected by $+11.6$ mean gain in the point-level metric).
Our contributions are as follows:
\begin{itemize}
    \item We propose a generative lane topology reasoning framework that models lane topology as a global structure and predicts lane graphs autoregressively.
    \item We leverage large-scale map data to learn the geometry prior over lane graphs, which can be transferred to benefit perception.
    \item We achieve substantial improvements over previous discriminative methods, producing spatially accurate and structurally realistic lane graphs.
\end{itemize}

\section{Related Work}
\noindent\textbf{Lane Topology Reasoning.}
Lane topology reasoning aims to detect lanes and infer their connectivity. Existing methods mainly fall into two paradigms.  
(1) DETR-style frameworks~\cite{STSU, CenterLineDet, TopoNet, TopoMLP,  ECFusion, Topo2D, LaneGAP, MV2D, CGNet, RoadPainter, TSTGT, TopoPoint, RATopo, li2026unified, TopoFormer} use learnable queries to detect lane instances and predict topological relations. Representative designs include interaction modeling by graph convolutional networks~\cite{TopoNet}, distance-to-relation mapping~\cite{TopoLogic}, and endpoint-aware reasoning~\cite{TopoPoint}.  
(2) Sequential graph generation methods~\cite{lu2023translating, lu2025translating, LaneGraph2Seq, SeqGrowGraph, Topo2Seq} formulate lane topology construction as autoregressive generation. They either serialize lanes into a token sequence and generate it with a sequence-to-sequence transformer~\cite{lu2023translating, lu2025translating}, or incrementally grow the graph by adding vertices or edges~\cite{SeqGrowGraph}.
Although these methods use sequence modeling, they focus on sophisticated graph traversal to determine the serialization order and predict lane sequences directly from perceptual inputs. In contrast, we focus on learning the geometry prior of lane graph structures from large-scale map data to facilitate lane topology reasoning.

\noindent\textbf{Map Perception with Prior.}
Many studies incorporate prior information to enhance online map perception for autonomous driving.
Most approaches derive priors from Standard Definition~(SD) maps~\cite{PMapNet, SMERF, MapLite2, SMART, Score, PriorDrive, SEPT}, satellite imagery~\cite{SMART, SatForHDMap}, and trajectories~\cite{TrajTopo}.
For instance, SMERF~\cite{SMERF} encodes SD maps as polyline sequences and fuses map priors via cross-attention. 
SatForHDMap~\cite{SatForHDMap} introduces a hierarchical fusion module that refines onboard sensor features by aligning them with satellite representations.
TrajTopo~\cite{TrajTopo} leverages crowdsourced trajectories, representing them as rasterized heatmaps and vectorized instance tokens, and incorporates them as priors for online mapping.
Unlike these methods that rely on extra inputs, we pre-train on large-scale map data to internalize a geometry prior into the model parameters.

\noindent\textbf{Generative Model for Map.}
Generative models have been explored for map construction. VectorMapNet~\cite{vectormapnet} uses a polyline generator to produce map elements. 
Several studies~\cite{DiffMap,MapPrior,DifFUSER,CoGMP,wang2024image} leverage sensor inputs within a diffusion framework to generate rasterized maps.
MapDiffusion~\cite{MapDiffusion} adopts diffusion to model the distribution of possible vectorized maps.
LaneDiffusion~\cite{LaneDiffusion} further employs diffusion models at the BEV feature level.
These generative methods help enforce global structural consistency in map perception.

\section{Method}
\label{sec:method}

\subsection{Overview}
\noindent\textbf{Problem Formulation.}~Given surround-view images captured by multiple cameras mounted on a vehicle, the lane topology reasoning task aims to detect centerlines \(\mathbf{L} = \{l_i \mid i = 1, 2, \dots, M\}\) in the BEV space and reason about their topology relationships \(\mathbf{E} \subseteq \mathbf{L} \times \mathbf{L}\).
Each centerline \(l_i\) is represented as a directed ordered sequence of points \(\mathbf{P} = [p_1, p_2, \dots, p_K]\), where \(p = (x, y) \in \mathbb{R}^2\) denotes a 2D coordinate in BEV space, and \(p_1\), \(p_{K}\) are the starting and ending points, respectively.
The topology relationship captures the connectivity among directed centerlines, forming a directed lane graph. An edge \((l_i, l_j) \in \mathbf{E}\) indicates that \(l_i\) is a predecessor of \(l_j\).

\begin{figure}[t]
  \centering
  \includegraphics[width=0.95\linewidth]{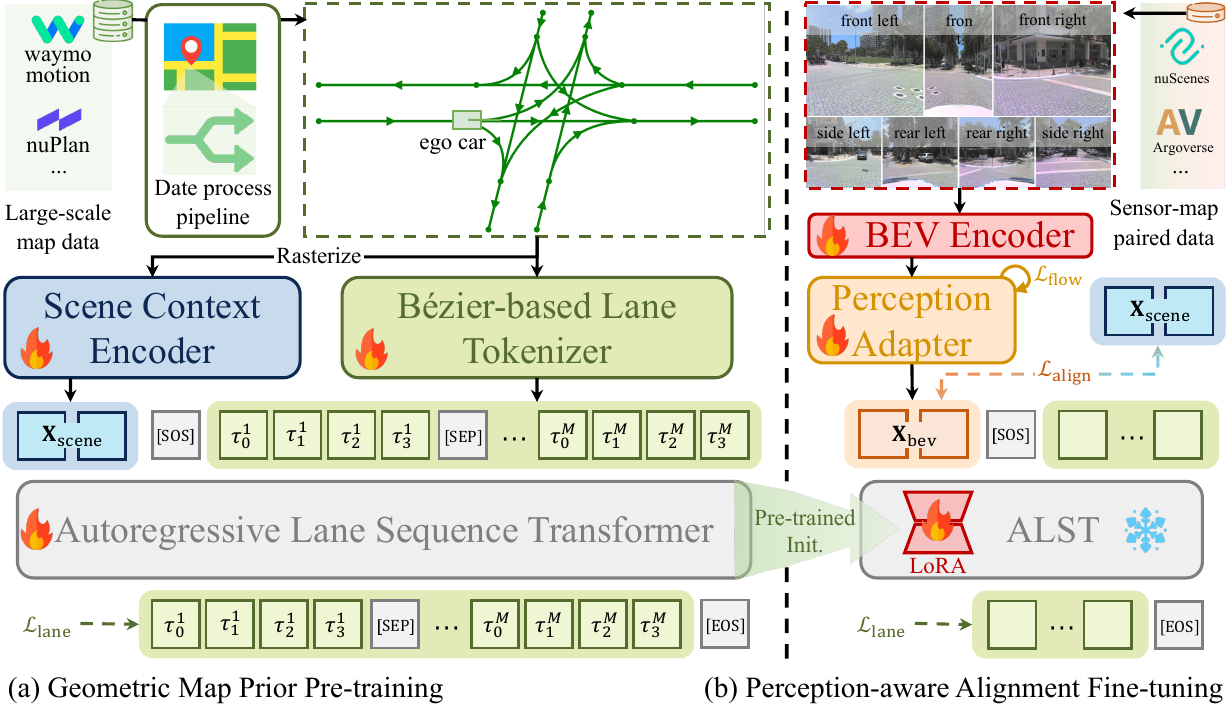}
  \caption{\textbf{Framework of TopoGPT.} Built on an autoregressive lane sequence transformer, we first perform geometric map-prior pre-training on large-scale map data to learn the geometry prior, where lane token sequences are generated conditioned on scene tokens \(\textbf{X}_{\text{scene}}\). We then conduct perception-aware alignment fine-tuning on sensor-map paired data, initializing from the pre-trained weights to leverage the learned prior, where a BEV encoder extracts BEV tokens \(\textbf{X}_{\text{bev}}\) from multi-view images and aligns them with \(\textbf{X}_{\text{scene}}\), enabling lane graph prediction from sensor data.}
  \label{fig:framework}
\end{figure}

\noindent\textbf{Framework Overview.}~
Unlike prior methods that decouple per-centerline detection from topology prediction, we note that lane graphs exhibit a strong geometry prior (\eg, parallel adjacent lanes and regular intersection connectivity). We therefore aim to explicitly model and leverage this geometry prior to produce accurate centerline positions and reasonable topological relationships.

To this end, we employ a generative model with an Autoregressive~(AR) architecture. 
Built upon this framework, we propose a two-stage training strategy, as illustrated in~\cref{fig:framework}. 
In the first stage, \emph{Geometric Map-Prior Pre-training}, we exploit scalable map-only data to learn the prior via autoregressive generative training. Concretely, each lane graph is simultaneously serialized into a lane token sequence and rasterized into a pseudo-image to extract scene tokens. The AR model is then trained to autoregressively generate the lane token sequence conditioned on scene tokens, capturing geometric regularities prevalent in real-world road scenes. Further details can be found in~\cref{sec:GMP}.
In the second stage, \emph{Perception-aware Alignment Fine-tuning}, we adapt the model to limited sensor-map paired data. Benefiting from the learned prior, fine-tuning reduces to a simpler mapping from perception features to the pre-trained condition space. Specifically, BEV features extracted from multi-view images are aligned with the scene tokens used during pre-training. The generative model then conditions on BEV features and produces lane graph predictions that are geometrically realistic and consistent with the perception input. We provide further details in~\cref{sec:PA}.
With this two-stage paradigm, we learn the geometry prior from real-world lane graphs using a generative model. When conditioned on perceptual input, the model exploits the prior to generate precise and geometrically realistic lane predictions, enabling accurate topology reasoning through simple distance-based endpoint association.

\subsection{Geometric Map-Prior Pre-training}
\label{sec:GMP}
To learn the geometry prior over lane graphs, we introduce \emph{Geometric Map-Prior Pre-training}, which casts lane topology prediction as an autoregressive sequence generation problem.
As illustrated in~\cref{fig:framework}~(a), we first extract high-quality lane graphs from large-scale HD map sources via a dedicated processing pipeline.
To bridge continuous geometry and discrete sequence, we propose a \textbf{Lane Tokenizer} that parameterizes each lane with \Bezier curves and converts it into an ordered sequence of discrete lane tokens. 
To provide global context for generation, a \textbf{Scene Context Encoder} rasterizes the vectorized lanes, extracts features, and maps them to scene tokens. 
Finally, a decoder-only \textbf{Autoregressive Model} conditions on the scene tokens as a prefix and learns the joint distribution over lane token sequences via standard next-token prediction.

\noindent\textbf{Data Collection.}~
With the large-scale deployment of vehicles equipped with advanced driver-assistance systems, a substantial corpus of HD map data has been accumulated. 
These map-only sources are not constrained by the availability of synchronized multi-view images, making them substantially easier to scale than sensor-map paired datasets.
We aim to learn the distribution over common real-world lane graph topologies from this corpus, which serves as a geometry prior for perception. 
Specifically, our data processing pipeline consists of three steps.
(1)~\textit{Scene Retrieval}. We obtain raw map data from large-scale driving-log datasets with HD map annotations (\eg, nuPlan~\cite{nuplan} and the Waymo Open Motion Dataset~\cite{waymo}). 
To reduce geographic redundancy, we subsample ego-poses within each log using a distance threshold.
Each sampled ego-pose defines the origin of a local coordinate, and we query centerline elements within a fixed radius from the HD map.
(2)~\textit{Instance Merging}. Different datasets adopt heterogeneous definitions for HD map elements. 
Centerline instances may be segmented by semantic transitions (\eg, changes in line-style attributes) or truncated at boundaries (\eg, stop lines). 
To unify the representation, we aggregate lane instances according to the predecessor-successor relationships. 
Following SLEDGE~\cite{SLEDGE}, we merge adjacent lanes along non-branching paths, effectively removing intermediate instances with a single predecessor and a single successor. 
After merging, lanes are split only at topology bifurcations such as merges and forks.
(3)~\textit{Geometric Sampling}. For each lane instance, we uniformly sample \(K\) points and use their \((x,y)\) coordinates as the geometric representation \(\mathbf{P} \in \mathbb{R}^{K \times 2}\).

\begin{figure}[t]
  \centering
  \includegraphics[width=0.95\linewidth]{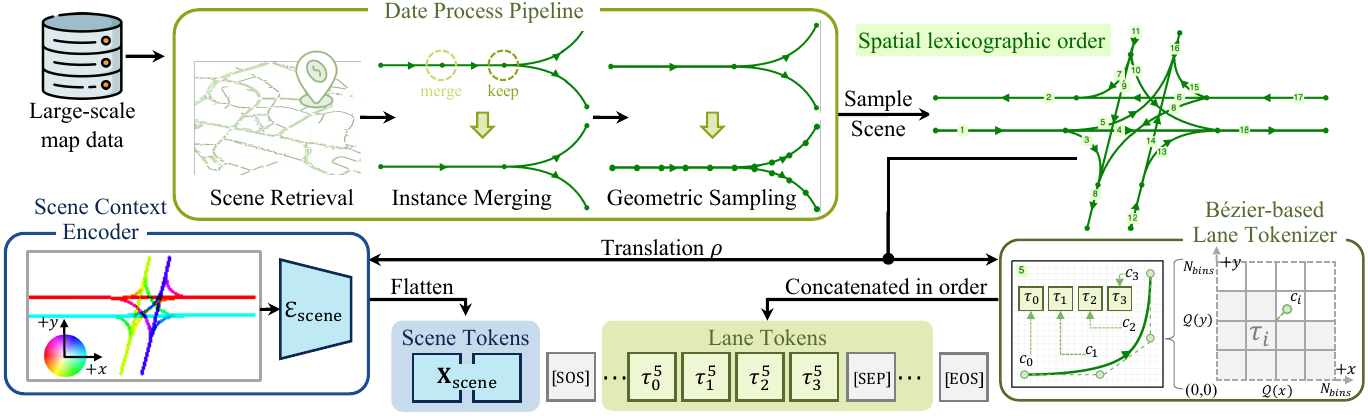}
  \caption{\textbf{Token sequence construction.} We first build a unified lane graph from large-scale map data through a data processing pipeline. For each sampled scene, we (1) rasterize it and encode with a scene context encoder to obtain scene tokens, (2) convert each lane into a discrete group using a \Bezier-based lane tokenizer and then sort all groups by a spatial lexicographical order to form the lane token sequence.}
  \label{fig:tokenizer}
\end{figure}

\noindent\textbf{Lane Tokenizer.}~To model the prior distribution of centerlines in map data with an AR model, we design a \Bezier-based Lane Tokenizer that converts the centerlines of a scene into discrete lane token representations as shown in~\cref{fig:tokenizer}.
To achieve a compact and unified representation of centerlines, for each centerline instance we fit a discrete coordinate sequence \(\mathbf{P}\) with a cubic \Bezier curve defined by four ordered control points \(\mathbb{C} = \{c_0, c_1, c_2, c_3\}\). This parameterization maps the discrete control points onto a continuous curve of the form:

\begin{equation}
B(t,\mathbb{C}) = \sum_{i=0}^{n} \binom{n}{i} t^i (1-t)^{n-i} c_i, \quad t \in [0,1],\; n=3.
\end{equation}
The control points are obtained by solving a least-squares problem that minimizes the discrepancy between the \(B(t,\mathbb{C})\) and \(\mathbf{P}\). 
The curve is strictly constrained to pass through the first and last control points, \ie, \(c_0=p_1\) and \(c_3=p_K\). In this formulation, \(c_0\) and \(c_3\) are directly assigned as the endpoints~(\ie, start/end points), ensuring that the global topological structure and connectivity are preserved. The intermediate control points \(c_1\) and \(c_2\) serve as shape descriptors that capture the local geometry of the lane.
We then quantize the continuous coordinates \((x, y)\) of each control point into discrete vocabulary indices. Inspired by Pix2seq~\cite{Pix2seq}, we define a quantization function \(\mathcal{Q}(v, v_{\max})\) that maps a continuous value \(v \in [0, v_{\max}]\) to an integer in the range \([1, N_{\text{bins}}]\):
\begin{equation}
\mathcal{Q}(v, v_{\max}) = \left\lfloor \frac{v}{v_{\max}} \cdot (N_{\text{bins}} - 1) \right\rfloor + 1.
\end{equation}
With this definition, we uniformly discretize both axes of the BEV range, with spatial extents \(X\) and \(Y\) respectively, into \(N_{\text{bins}}\) bins, so that four control points \(\mathbb{C}\) can be represented as a sequence of discrete token indices as:
\begin{equation}
\mathbf{t} = \left\{ \tau_i \mid \tau_i = \mathcal{Q}(y_i, Y) \cdot N_{\text{bins}} + \mathcal{Q}(x_i, X), \quad i \in \{0, 1, 2, 3\} \right\}.
\end{equation}

Furthermore, to establish a deterministic ordering among centerlines, we apply a spatial lexicographic sort~\cite{ScenarioDreamer} that prioritizes the \([x_{\min}, y_{\min}, x_{\max}, y_{\max}]\) coordinates sequentially. This spatial sorting organizes lanes in canonical left-to-right (and then bottom-to-top) order.
The discretized token group \(\mathbf{t}^j\) of each centerline \(l_j\) is then arranged according to this order. 
We insert a separator token \( \texttt{<SEP>} \) between adjacent centerlines in the sorted order, and prepend/append special tokens \( \texttt{<SOS>} \) and \( \texttt{<EOS>} \) to mark the beginning and end of the sequence, respectively.
Through the above process, the centerlines can be converted into a discrete token sequence as:
\begin{equation}
\mathbf{T} = [ \texttt{<SOS>} ,\; \ldots,\;  \texttt{<SEP>} ,\; \underbrace{\tau_0^j, \tau_1^j, \tau_2^j, \tau_3^j}_{\mathbf{t}^j},\;\ldots,\;  \texttt{<SEP>} ,\; \underbrace{\tau_0^M, \tau_1^M, \tau_2^M, \tau_3^M}_{\mathbf{t}^M},\;  \texttt{<EOS>} ],
\end{equation}
where \(M\) denotes the number of centerlines in the scene, and \(\tau_i^j\) represents the quantized token of the \(i\)-th control point of the \(j\)-th centerline.

\noindent\textbf{Scene Context Encoder.}~Unconditional generation of complex lane topologies lacks explicit spatial constraints, making the model prone to drifting away from the scene distribution. To mitigate this issue, we introduce globally informative scene tokens as the condition.
Motivated by the map encoder used in motion planners~\cite{SLEDGE}, we define a vector-to-raster translation \(\rho\) that converts the raw vectorized centerline set \(\mathbf{L} \in \mathbb{R}^{M \times K \times 2}\) into a rasterized scene image \(\mathbf{I}_l \in \mathbb{R}^{X_i \times Y_i \times 2}\). 
Specifically, for each pixel traversed by a polyline segment, we assign a two-dimensional direction vector \(\Delta = [dx, dy]\) pointing to its successor point, while all remaining pixels are set to background values (\ie, \([0, 0]\)) as in~\cref{fig:tokenizer}. This translation preserves geometric orientation and topological connectivity of the lane graph.
To extract compact features from the high-resolution rasterized image, we use a convolutional neural network encoder \(\mathcal{E}_{\text{scene}}\), which downsamples \(\mathbf{I}_l\) into a feature map \(\mathbf{F}_{\text{scene}} \in \mathbb{R}^{X_f \times Y_f \times C}\). 
To bridge the modality gap between raster features and discrete tokens, we follow a standard vision-language alignment pipeline: we flatten \(\mathbf{F}_{\text{scene}}\) along spatial dimensions and project it into the AR model hidden dimension \(D\) using a linear layer, followed by LayerNorm to stabilize feature scaling. 
The projected features form a scene token sequence \(\mathbf{X}_{\text{scene}} \in \mathbb{R}^{N_{\text{scene}} \times D}\), which serve as prefix conditions, guiding the AR model to sequentially predict lane tokens \(\mathbf{T}\). This process is formulated as:
\begin{equation}
\mathbf{X}_{\text{scene}}
= \mathrm{LN}\Bigl(
    \mathrm{Proj}\Bigl(
        \mathrm{Flatten}\bigl(
            \mathcal{E}_{\text{scene}}\bigl(
                \rho(\mathbf{L})
            \bigr)
        \bigr)
    \Bigr)
\Bigr).
\end{equation}

\noindent\textbf{Autoregressive Model.}~We adopt an AR architecture similar to LLaMAGen~\cite{LLamaGen}. Specifically, we reformulate structured lane topology reasoning as a sequence modeling problem and propose the Autoregressive Lane Sequence Transformer (ALST). ALST is based on a decoder-only Transformer~\cite{Transformer} \(\mathcal{G}_\theta\) with a standard causal mask. On the input side, the scene-token sequence \(\mathbf{X}_{\text{scene}}\) is used as a prefix condition to provide scene-level cues for lane token generation, \ie, \(\mathbf{T} = \mathcal{G}_\theta(\mathbf{X}_{\text{scene}})\). Notably, we adopt Multimodal Rotary Position Embedding (M-RoPE) from Qwen2-VL~\cite{qwen2} for scene tokens to preserve 2D spatial locality after flattening.
During generation, the model predicts each next lane token conditioned on the scene context and all previously generated lane tokens. Formally, the conditional probability of the sequence \(\mathbf{T}\) is factorized as:

\begin{equation}
p(\mathbf{T} \mid \mathbf{X}_{\text{scene}}) = \prod_{m=1}^{|\mathbf{T}|} p(\tau_m \mid \mathbf{X}_{\text{scene}},\; \tau_{<m}).
\end{equation}

During training, we adopt a teacher forcing strategy and optimize both the scene context encoder \(\mathcal{E}_{\text{scene}}\) and the AR decoder \(\mathcal{G}_\theta\) end-to-end by minimizing the Cross-Entropy (CE) loss \(\mathcal{L}_{\text{lane}}\) over the predicted lane token sequence.

\noindent\textbf{Training Strategy.}~During pre-training, we aim for the model to not only generate scene-consistent centerlines, but also capture intrinsic geometry prior reflected in lane token sequences. To this end, we apply the following data augmentations.
(1) \textit{Global Augmentation.} To prevent overfitting and improve generalization, we apply the same geometric transformation to both the rasterized scene image and the vectorized centerlines, including random flipping, translation, and rotation.
(2) \textit{Condition Degradation.} To prevent the model from over-relying on the scene condition and neglecting inherent geometric structure, we randomly degrade the scene condition during training. Specifically, we randomly remove a subset of lanes from the rasterized scene image, or drop the scene condition entirely for some samples, while always keeping the full information in the lane token sequences. This completion objective trains the model to predict the full lane token sequence under degraded scene conditioning, thereby encouraging it to learn the lane graph topology.

\subsection{Perception-aware Alignment Fine-tuning}
\label{sec:PA}
\begin{figure}[t]
  \centering
  \includegraphics[width=0.7\linewidth]{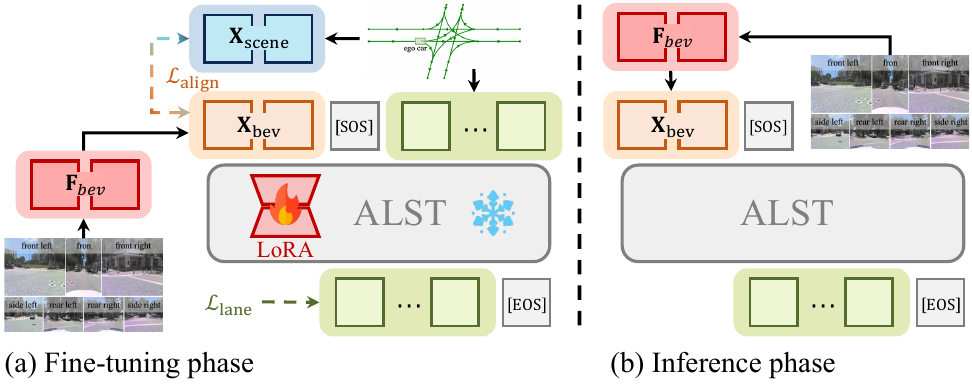}
  \caption{(a) During fine-tuning, \(\mathbf{X}_{\text{bev}}\) is optimized under alignment and lane prediction objectives. (b) During inference, \(\mathbf{X}_{\text{bev}}\) is directly used as condition for lane prediction.}
  \label{fig:finetune}
\end{figure}

To leverage the learned geometry prior for lane prediction from sensory inputs, we extract perceptual features through a \textbf{BEV Encoder} and align them with scene tokens used during pre-training via a \textbf{Perception Adapter} as shown in~\cref{fig:finetune}, thereby enabling lane prediction conditioned on multi-view images.

\noindent\textbf{BEV Encoder.}~Given multi-view images as input, we first employ a ResNet-50~\cite{resnet} backbone to extract multi-scale fused 2D feature maps from each view. Following~\cite{simpleBEV}, we adopt a bilinear sampling strategy for 2D-to-3D feature lifting. Specifically, a regular coordinate grid, which is spatially aligned with the rasterized scene image, is predefined in the target 3D space, and these 3D coordinates are back-projected onto the 2D feature maps of all cameras. Bilinear interpolation is performed at the projected locations to retrieve local image features, which are subsequently aggregated via weighted averaging to construct a unified 3D feature volume. The height dimension is then collapsed and folded into the channel dimension, yielding the BEV feature map \(\mathbf{F}_{\text{bev}} \in \mathbb{R}^{X_f \times Y_f \times C}\).

\noindent\textbf{Perception Adapter.} We design a Perception Adapter based on noise-free flow matching to map the BEV features into the scene tokens learned during pre-training.
Flow matching is a general generative modeling framework that learns a continuous transport trajectory from a source distribution to a target distribution by solving the ordinary differential equation \(\frac{dz_t}{dt} = v_\theta(z_t, t)\), where \(z_t\) denotes the evolving sample at continuous time \(t \in [0, 1]\) and \(v_\theta\) is a neural network parameterized velocity field. 
Recent work~\cite{CrossFlow, flowtok} has demonstrated that, unlike diffusion models that typically assume a Gaussian source distribution, flow matching does not theoretically require a specific parametric form for source distribution. This flexibility enables the model to start from structured feature distributions and learn a direct transport path to the target distribution.

Building on this property, we propose a flow-matching-based adapter that evolves BEV features toward the pre-trained scene token distribution. We formulate the procedure as a cross-modality transport problem: the continuous BEV features extracted from multi-view images define the source distribution, while the target distribution is given by the scene tokens produced by a frozen pre-trained scene context encoder.
Within the adapter, the \(v_\theta\), implemented as a lightweight DiT~\cite{DiT} model, is trained to approximate the ground-truth velocity that connects the source and target distributions along a linear interpolation path. Specifically, we optimize the flow-matching loss \(\mathcal{L}_{\text{flow}}\), defined as the Mean Squared Error (MSE) between the predicted velocity and the ground-truth velocity.
To mitigate the distribution shift in ALST conditioning from scene tokens during pre-training to BEV tokens during fine-tuning, we execute the flow-matching generation process during fine-tuning to obtain BEV-conditioned tokens.
We further introduce a latent alignment loss \(\mathcal{L}_{\text{align}}\) to supervise the adapter. During each forward pass, we explicitly unroll the numerical integration steps starting from the \(\mathbf{F}_{\text{bev}}\) to obtain the predicted BEV tokens \(\mathbf{X}_{\text{bev}} \in \mathbb{R}^{N_{\text{scene}} \times D} \), which are aligned with the corresponding scene tokens using a feature-level MSE loss \(\mathcal{L}_{\text{align}}\).
Moreover, \(\mathbf{X}_{\text{bev}}\) is simultaneously fed to the subsequent ALST as the prefix condition. The CE loss \(\mathcal{L}_{\text{lane}}\), identical to that used in pre-training, is computed on the predicted lane token sequence and backpropagated through the unrolled integration steps to update the adapter end-to-end.

\noindent\textbf{Training.}~We jointly optimize the BEV encoder and the Perception Adapter, while employing Low-Rank Adaptation (LoRA) to fine-tune the pre-trained ALST so that it adapts to the BEV token inputs.

\noindent\textbf{Inference.}~During inference, to ensure stable predictions, we start with the BEV features \(\mathbf{F}_{\text{bev}}\) and obtain the BEV tokens \(\mathbf{X}_{\text{bev}}\) through a deterministic ODE integration process (implemented by an Euler solver with discretization steps \(N_\text{flow}\)). Moreover, we use greedy decoding for autoregressive generation. To guarantee that the lane token sequence can be correctly decoded, we perform constrained decoding by forcing the \texttt{<SEP>} token at fixed periodic positions. For topology prediction, benefiting from the high geometric consistency at the endpoints of our predicted centerlines, we simply apply a distance threshold \(\epsilon = 0.5\,\text{m}\) at endpoints to determine predecessor-successor relationships.

\section{Experiments}
\label{sec:exp}
\subsection{Experimental Setup}

\noindent\textbf{Datasets.}~For the pre-training dataset, we extract HD map information from motion datasets including Waymo Motion~\cite{waymo}, nuPlan~\cite{nuplan}, and Argoverse 2 Motion~\cite{av2}, constructing $2.07$M, $1.01$M, and $0.25$M lane graph scenes respectively, yielding a total of $3.3$M samples for pre-training.
For the fine-tuning dataset, we evaluate our method on the OpenLane-V2~\cite{Openlanev2}, which is divided into two subsets: \textit{subset\_A} and \textit{subset\_B}, each containing 1,000 scenes captured at 2 Hz with multi-view images and corresponding annotations. We utilize its annotations for lane centerline detection and inter-lane topology reasoning tasks.

\noindent\textbf{Methods under Comparison.}~We select several recent DETR-based state-of-the-art methods for comparison, including TopoNet~\cite{TopoNet}, TopoMLP~\cite{TopoMLP}, TopoLogic~\cite{TopoLogic}, and TopoPoint~\cite{TopoPoint}. Notably, these methods typically output far more predictions than the actual number of lanes in a scene (\eg, $200$ or $300$ candidates) along with associated confidence scores, whereas our method directly predicts all lanes in the scene. To ensure a fair comparison, we analyze the precision-recall curve as shown in~\cref{fig:pr_curve} for each competing method and identify the operating point that maximizes the F1-score, using the corresponding predictions as the optimal output for that method.

\noindent\textbf{Evaluation Metric.}~Based on the predicted centerline graph and the Ground Truth~(GT), we employ two categories of evaluation metrics. (1)~\textit{Lane-level metrics}: adapted from \cite{Openlanev2}, we report the Mean Precision (M-P) and Mean Recall (M-R) of centerline averaged over multiple distance thresholds, along with the Mean Topology (M-T) similarity between the predicted and the GT. (2)~\textit{Point-level metrics}: Following~\cite{LaneGNN, Aerial_Image2Map, LaneGAP, CGNet}, we convert the predicted centerlines and their inter-lane topology into a point-level graph. The structural similarity is then evaluated across two key dimensions: spatial alignment via Graph IoU (G-IoU), Split Detection Accuracy (SDA), and GEO F1 (G-F1), while structural fidelity is assessed using TOPO F1 (T-F1), JTOPO F1 (JT-F1), and Average Path Length Similarity (APLS).

\noindent\textbf{Implementation Details.}~In the pre-training stage, we set the BEV range to \([\pm 52\,\text{m}, \pm 32\,\text{m}]\) along the X and Y axes respectively, and adopt \(N_{\text{bins}} = 64\) to discretize the coordinate grid. The model is trained for $16$ epochs on $8$ NVIDIA H20 GPUs with a batch size of $128$ per GPU. In the fine-tuning stage, we follow \cite{TopoNet} and apply the same image-level data augmentation strategy. The model is trained for $24$ epochs on $8$ NVIDIA H20 GPUs with a batch size of $8$ per GPU.

\begin{table}[t]
    \centering
    \caption{Performance comparison on OpenLane-V2. We use their released model checkpoints when available; otherwise, we reproduce the results using the official codebases and adapt the outputs to our metric definitions.}
    \label{tab:main_result}
    % Adjust this value to control horizontal spacing between columns
    \setlength{\tabcolsep}{3pt}
    \begin{tabular}{c | l | ccc | cccccc}
        \toprule
        \multirow{2}{*}{Data} & \multirow{2}{*}{Method} & \multicolumn{3}{c|}{lane-level metric$\uparrow$} & \multicolumn{6}{c}{point-level metric$\uparrow$} \\
        & & M-P & M-R & M-T & G-IoU & SDA & G-F1 & T-F1 & JT-F1 & APLS \\
        \hline
        
        % Subset A data section
        \multirow{5}{*}{\textit{A}}
        & TopoNet~\cite{TopoNet}      & 39.7 & 32.3 & 10.1 & 38.8 & 7.40  & 45.6 & 32.3 & 21.7 & 9.80  \\
        & TopoMLP~\cite{TopoMLP}      & 40.4 & 33.1 & 13.9 & 40.5 & 11.3 & 47.5 & 34.4 & 24.4 & 11.4 \\
        & TopoLogic~\cite{TopoLogic}    & 40.5 & 33.6 & 15.5 & 39.0 & 11.5 & 45.8 & 32.0 & 21.7 & 10.4 \\
        & TopoPoint~\cite{TopoPoint}    & 42.7 & 34.0 & 16.6 & 40.6 & 19.6 & 47.5 & 34.5 & 24.2 & 17.4 \\
        & TopoGPT (Our) & \textbf{45.6} & \textbf{42.6} & \textbf{24.4} & \textbf{54.4} & \textbf{26.9} & \textbf{60.3} & \textbf{47.4} & \textbf{35.7} & \textbf{28.7} \\
        \hline
        
        % Subset B data section
        \multirow{5}{*}{\textit{B}}
        & TopoNet~\cite{TopoNet}      & 33.4 & 26.3 & 6.27 & 31.5 & 4.20  & 36.3 & 24.3 & 18.1 & 6.40  \\
        & TopoMLP~\cite{TopoMLP}      & 36.8 & 28.3 & 12.5 & 39.7 & 9.10  & 42.3 & 29.7 & 23.1 & 12.2 \\
        & TopoLogic~\cite{TopoLogic}    & 33.9 & 27.2 & 11.7 & 36.7 & 8.60  & 40.1 & 28.0 & 21.4 & 10.3 \\
        & TopoPoint~\cite{TopoPoint}    & 35.4 & 26.8 & 12.6 & 31.9 & 13.2 & 36.2 & 25.7 & 19.6 & 11.1 \\
        & TopoGPT (Our) & \textbf{43.1} & \textbf{39.2} & \textbf{23.4} & \textbf{55.4} & \textbf{25.8} & \textbf{58.9} & \textbf{46.3} & \textbf{36.8} & \textbf{29.5} \\
        \bottomrule
    \end{tabular}
\end{table}

\subsection{Quantitative Evaluation}
As shown in~\cref{tab:main_result}, our method substantially outperforms existing approaches across multiple metrics on both subsets. On \textit{subset\_A}, regarding lane-level metrics, \ourmethod achieves an average $+6.4$ improvement, with more accurate centerline localization and shape prediction compared to the previous best results ($\textbf{45.6}$ vs. $42.7$ on M-P, $\textbf{42.6}$ vs. $34.0$ on M-R) and more reasonable topology relationship ($\textbf{24.4}$ vs. $16.6$ on M-T). The notable $+8.6$ improvement on Recall further demonstrates that the learned geometry prior can reasonably complete the lane graph. Regarding point-level metrics, \ourmethod achieves an average $+11.6$ improvement, with more precise geometric structure prediction ($\textbf{54.4}$ vs. $40.6$ on G-IoU, $\textbf{26.9}$ vs. $19.6$ on SDA, $\textbf{60.3}$ vs. $47.5$ on G-F1) and more accurate connectivity inference ($\textbf{47.4}$ vs. $34.5$ on T-F1, $\textbf{35.7}$ vs. $24.4$ on JT-F1, $\textbf{28.7}$ vs. $17.4$ on APLS). On \textit{subset\_B}, the improvements are equally significant.

\subsection{Ablation Study}
We conduct ablation studies on \textit{subset\_A} and primarily report lane-level metrics.

\begin{figure}[t]
    \centering
    \parbox{0.49\linewidth}{
        \centering
        \includegraphics[width=\linewidth]{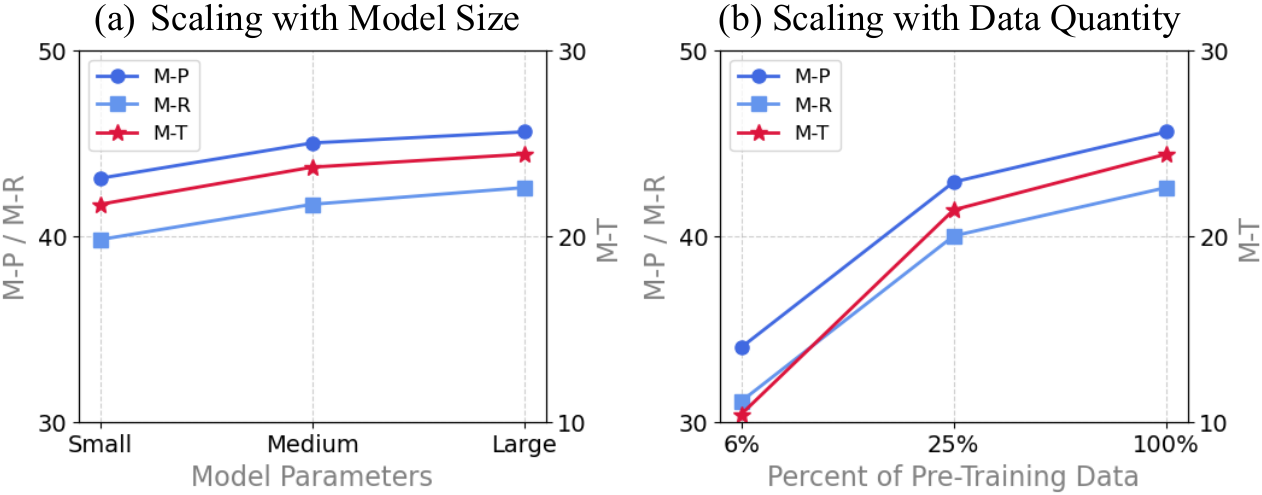}
        \caption{Performance of the fine-tuned models under different pre-training model sizes and data quantities. As the number of model parameters and the percentage of pre-training data increase, performance consistently improves across all metrics.}
        \label{fig:scaling}
    }
    \hfill
    \parbox{0.47\linewidth}{
        \centering
        \includegraphics[width=\linewidth]{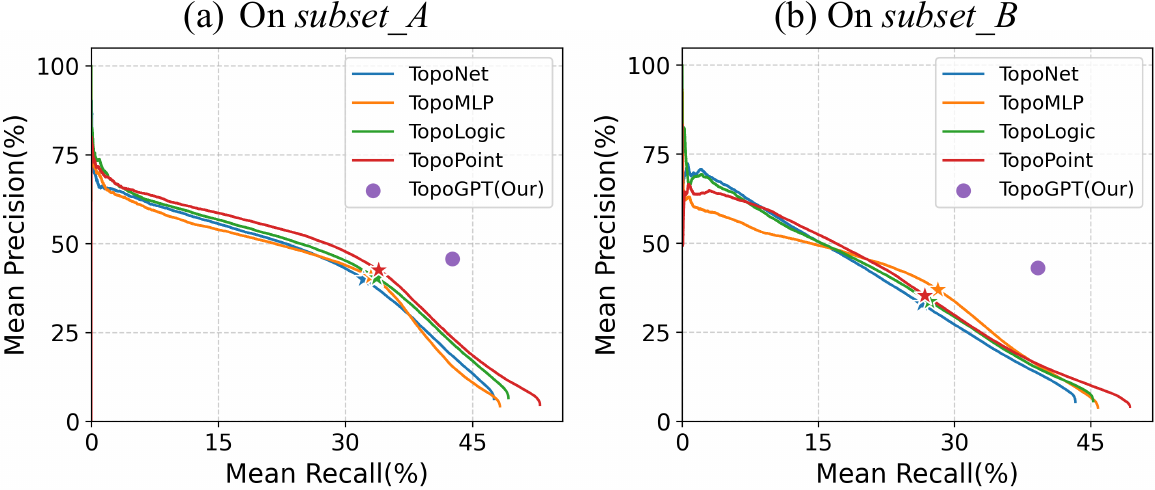}
        \caption{Precision-Recall curve for lane-level metrics: we plot the Mean Precision \vs Mean Recall as the confidence threshold varies. Stars indicate the best F1-score operating points of previous methods, while the circle marks our method.}
        \label{fig:pr_curve}
    }
\end{figure}

\noindent\textbf{Scalability of pre-training.} During pre-training, \ourmethod demonstrates favorable scalability with respect to both model size and data quantity. (a) \textit{Model size}:~\cref{fig:scaling}~(a) illustrates that as the model scales from Small ($24.7$M) to Medium ($45.9$M) to Large ($91.3$M) parameters, all evaluation metrics improve consistently. (b) \textit{Data quantity}: Based on the TopoGPT-L model, the results in~\cref{fig:scaling}~(b) reveal that increasing the amount of pre-training data leads to steady performance gains of the final fine-tuned model, indicating that the geometry prior acquired during pre-training effectively transfers to and benefits the downstream fine-tuning stage.

\noindent\textbf{Training Strategy in pre-training.} \cref{tab:pretrain-aug} validates the effectiveness of the data augmentation strategies introduced in the pre-training stage. Incorporating Global Augmentation (G.A.) and Condition Degradation (C.D.) each independently enhances the model's ability to learn geometry prior, thereby improving various metrics after fine-tuning. Moreover, the two strategies exhibit a synergistic effect, jointly contributing $+4.6$/$+4.2$ improvements on M-P/M-R and a $+3.3$ gain on M-T.

\begin{table}[t!]
    \centering
    % Left parbox (adjust 0.48 to control width ratio)
    \parbox{0.48\linewidth}{
        \centering
        \caption{Ablation study on pre-training strategies, where \textbf{G.A.}=Global Augmentation, \textbf{C.D.}=Condition Degradation.}
        \footnotesize % Shrink font size
        \begin{tabular*}{\linewidth}{@{\extracolsep{\fill}} cc | ccc @{}}
            \toprule
            \textbf{G.A.} & \textbf{C.D.} & M-P$\uparrow$ & M-R$\uparrow$ & M-T$\uparrow$ \\
            \hline
            &  & 41.0 & 38.4 & 21.1 \\
            \checkmark & & 42.6 & 40.1 & 22.1 \\
            & \checkmark & 43.3 & 40.8 & 22.6 \\
            \checkmark & \checkmark & \textbf{45.6} & \textbf{42.6} & \textbf{24.4} \\
            \bottomrule
        \end{tabular*}
        \label{tab:pretrain-aug}
    }
    \hfill % Fill horizontal space between boxes
    % Right parbox
    \parbox{0.48\linewidth}{
        \centering
        \caption{Ablation study on fine-tuning schemes.}
        \footnotesize
        \begin{tabular*}{\linewidth}{@{\extracolsep{\fill}} l | ccc @{}}
            \toprule
             & M-P$\uparrow$ & M-R$\uparrow$ & M-T$\uparrow$ \\
            \hline
            From Scratch   & 7.4 & 7.0 & 0.8 \\
            Freeze & 44.3 & 42.1 & 23.7 \\
            LoRA   & \textbf{45.6} & \textbf{42.6} & \textbf{24.4} \\
            Full Fine-tune & 45.3 & 42.2 & 24.3 \\
            \bottomrule
        \end{tabular*}
        \label{tab:finetune-type}
    }
\end{table}

\noindent\textbf{Training Strategy in fine-tuning.} We investigate how to leverage the pre-trained ALST weights during the fine-tuning stage. As shown in~\cref{tab:finetune-type}, \textit{From Scratch} refers to training ALST to predict lane token sequences from BEV features without inheriting pre-trained weights. In this setting, the model struggles to learn such a complex mapping from limited sensor-map paired data, resulting in notably poor results. When inheriting the pre-trained weights, we explore three adaptation schemes: freezing all parameters, LoRA-based fine-tuning, and full-parameter fine-tuning, all of which achieve comparably high performance. We adopt LoRA by default due to its parameter efficiency. Overall, incorporating pre-training yields improvements of $+38.2$/$+35.6$/$+23.6$ on M-P/M-R/M-T, confirming the effectiveness of the geometry prior learned from pre-training.

\begin{table}[t!]
    \centering
    % Left parbox (adjust 0.48 to control width ratio)
    \parbox{0.6\linewidth}{
        \centering
        \caption{Ablation study on fine-tuning loss designs. \texttt{w/o} flow denotes directly using the output features from adapter.}
        \footnotesize % Shrink font size
        \begin{tabular*}{\linewidth}{@{\extracolsep{\fill}} c | cc | ccc @{}}
            \toprule
            & $\mathcal{L}_{align}$ & $\mathcal{L}_{lane}$ & M-P$\uparrow$ & M-R$\uparrow$ & M-T$\uparrow$ \\
            \hline
            \texttt{w/o} flow & \checkmark & \checkmark & 43.1 & 39.6 & 21.6\\
            \hline
            \multirow{3}{*}{\texttt{w/} flow}& \checkmark & & 42.0 & 39.5 & 21.7 \\
            & & \checkmark & 37.2 & 37.0 & 19.6 \\
            & \checkmark & \checkmark & \textbf{45.6} & \textbf{42.6} & \textbf{24.4} \\
            \bottomrule
        \end{tabular*}
        \label{tab:finetune-loss}
    }
    \hfill % Fill horizontal space between boxes
    % Right parbox
    \parbox{0.35\linewidth}{
        \centering
        \caption{Ablation study on flow-matching steps.}
        \footnotesize
        \begin{tabular*}{\linewidth}{@{\extracolsep{\fill}} c | ccc @{}}
            \toprule
             $N_{\text{flow}}$ & M-P$\uparrow$ & M-R$\uparrow$ & M-T$\uparrow$ \\
            \hline
            0 & 43.1 & 39.6 & 21.6 \\
            3 & 44.3 & 41.0 & 22.8 \\
            6 & \textbf{45.6} & \textbf{42.6} & \textbf{24.4} \\
            9 & 45.4 & 41.5 & 23.4 \\
            \bottomrule
        \end{tabular*}
        \label{tab:finetune-flow-step}
    }
\end{table}

\noindent\textbf{Loss Terms in fine-tuning.} \cref{tab:finetune-loss} presents the contribution of each loss term during the fine-tuning. We first establish a baseline adapter without flow matching, where the output features of the DiT model are directly used as \(\mathbf{X}_{\text{bev}}\) and aligned with \(\mathbf{X}_{\text{scene}}\). This approach already achieves competitive performance, and introducing flow matching to model the transition between the two token distributions further brings $+2.5$/$+3.0$/$+2.8$ improvements on M-P/M-R/M-T. Meanwhile, both \(\mathcal{L}_{\text{align}}\) and \(\mathcal{L}_{\text{lane}}\) serve critical roles. In particular, the notable accuracy drop upon removing \(\mathcal{L}_{\text{lane}}\) underscores the importance of end-to-end optimizing BEV tokens through the lane token prediction loss.

\noindent\textbf{Integration Steps in fine-tuning.} \cref{tab:finetune-flow-step} shows that the performance remains consistent across different integration steps \(N_{\text{flow}}\). This validates the effectiveness of the flow-based perception adapter.

\subsection{Qualitative Results}

\begin{figure}[t]
  \centering
  \includegraphics[width=1.0\linewidth]{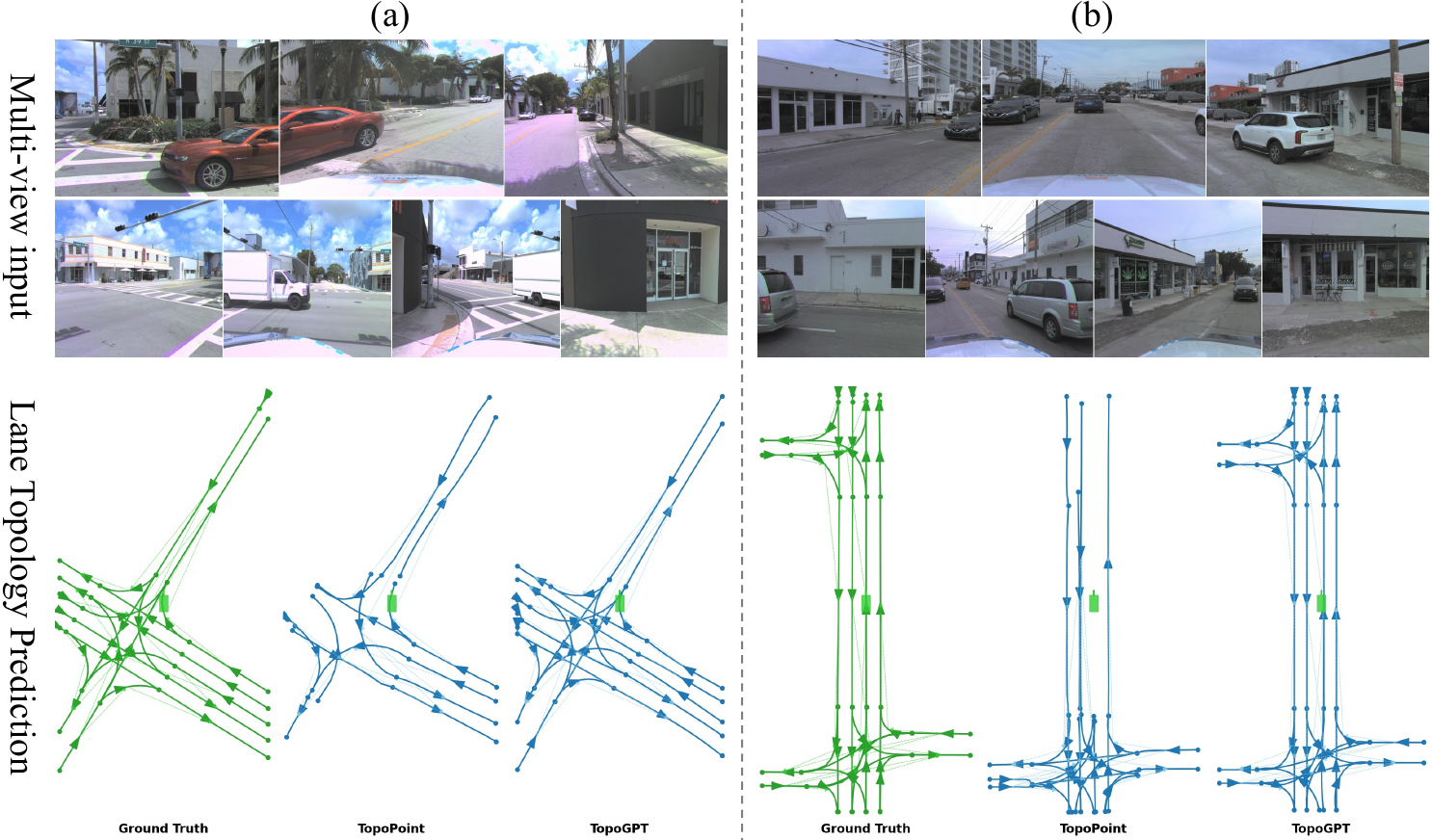}
  \caption{Qualitative comparison between TopoPoint and our \ourmethod. The first row shows multi-view inputs, and the second row denotes the lane graphs. Dark solid arrows indicate centerlines, while light dashed arrows depict topological relations between centerlines.}
  \label{fig:result}
\end{figure}

We present a qualitative comparison between TopoPoint and TopoGPT in~\cref{fig:result}. Specifically, we select two traffic scenarios for analysis and visualize the results of centerline detection and topology reasoning. In~\cref{fig:result}~(a), for a complex intersection, our \ourmethod produces a structurally complete and reasonably connected topology, whereas TopoPoint fails to predict necessary opposing and connecting centerlines. Meanwhile, the predictions of \ourmethod exhibit better geometric consistency at lane endpoints. In~\cref{fig:result}~(b), due to long-range perception limitations and partial occlusion, TopoPoint fails to detect the intersection directly ahead, while our \ourmethod is able to infer the centerlines and their connectivity at the intersection based on visual cues and existing lane structures. In summary, by leveraging geometry prior from large-scale map data, \ourmethod is capable of predicting complete and coherent lane graphs for entire scenes under the guidance of visual perception.

\section{Conclusion}
In this paper, we present TopoGPT, a generative framework for lane topology reasoning that models lane graphs via autoregressive sequence prediction. Pre-training on $3.3$M scenes of large-scale map data enables the model to learn the geometry prior of lane graphs, which are effectively transferred to perception-conditioned prediction via a flow-based perception adapter. Experiments on OpenLane-V2 demonstrate that TopoGPT substantially outperforms existing discriminative methods on both lane-level and point-level metrics, yielding lane graphs with improved geometric consistency and structural completeness.

\noindent\textbf{Limitations.}~Our method relies on autoregressive generation, which introduces sequential decoding latency compared to parallel detection methods. Moreover, autoregressive decoding inherently suffers from the risk of error accumulation, where inaccuracies in earlier predictions may propagate to subsequent lane generations due to sequential dependencies. Although global scene context helps mitigate this issue, it cannot be completely eliminated in complex or highly occluded scenarios. Additionally, the geometry prior learned during pre-training may limit generalization to regions with uncommon road layouts.

% \clearpage  % TODO FINAL: This \clearpage needs to be removed from both review and camera-ready versions.

\section*{Acknowledgements}
This research is supported in part by the Key Research Program of Hangzhou (No.~2025SZD1A56), the National Natural Science Foundation of China (No.~6246\-1160308, U23B2010, and 62576024), the Beijing Natural Science Foundation (No.~L231011), the Fundamental Research Funds for the Central Universities (No.~501RCQD2025141003), BeiHang GanWei Project (No.~502GWXM20241410\-01), the National Science Foundation Support Projects (No.~62425303), the National Key R\&D Program of China (No.~2024YFb4707300), and Young Elite Scientist Sponsorship Program by Beijing Association for Science and Technology.

\bibliographystyle{splncs04}
\bibliography{main}

@String(CVPR  = {IEEE Conf. Comput. Vis. Pattern Recog.})

@String(ICCV  = {Int. Conf. Comput. Vis.})

@String(ECCV  = {Eur. Conf. Comput. Vis.})

@String(NeurIPS = {Adv. Neural Inform. Process. Syst.})

@String(ICML  = {Int. Conf. Mach. Learn.})

@String(ICLR  = {Int. Conf. Learn. Represent.})

@String(AAAI  = {AAAI})

@String(TCSVT = {IEEE Trans. Circuit Syst. Video Technol.})

@String(CVPR  = {CVPR})

@String(ICCV  = {ICCV})

@String(ECCV  = {ECCV})

@String(NeurIPS = {NeurIPS})

@String(ICML  = {ICML})

@String(ICLR  = {ICLR})

@String(TCSVT = {IEEE TCSVT})

@inproceedings{MapPrior,
  author       = {Xiyue Zhu and
                  Vlas Zyrianov and
                  Zhijian Liu and
                  Shenlong Wang},
  title        = {MapPrior: Bird's-Eye View Map Layout Estimation with Generative Models},
  booktitle    = {{ICCV}},
  year         = {2023}
}

@inproceedings{Openlanev2,
  author       = {Huijie Wang and
                  Tianyu Li and
                  Yang Li and
                  Li Chen and
                  Chonghao Sima and
                  Zhenbo Liu and
                  Bangjun Wang and
                  Peijin Jia and
                  Yuting Wang and
                  Shengyin Jiang and
                  Feng Wen and
                  Hang Xu and
                  Ping Luo and
                  Junchi Yan and
                  Wei Zhang and
                  Hongyang Li},
  title        = {OpenLane-V2: {A} Topology Reasoning Benchmark for Unified 3D {HD}
                  Mapping},
  booktitle    = {NeurIPS},
  year         = {2023}
}

@article{PMapNet,
  author       = {Zhou Jiang and
                  Zhenxin Zhu and
                  Pengfei Li and
                  Huan{-}ang Gao and
                  Tianyuan Yuan and
                  Yongliang Shi and
                  Hang Zhao and
                  Hao Zhao},
  title        = {P-MapNet: Far-Seeing Map Generator Enhanced by Both SDMap and HDMap
                  Priors},
  journal      = {IEEE RAL},
  year         = {2024}
}

@inproceedings{Pix2seq,
  author       = {Ting Chen and
                  Saurabh Saxena and
                  Lala Li and
                  David J. Fleet and
                  Geoffrey E. Hinton},
  title        = {Pix2seq: {A} Language Modeling Framework for Object Detection},
  booktitle    = {{ICLR}},
  year         = {2022}
}

@inproceedings{ScenarioDreamer,
  author       = {Luke Rowe and
                  Roger Girgis and
                  Anthony Gosselin and
                  Liam Paull and
                  Christopher Pal and
                  Felix Heide},
  title        = {Scenario Dreamer: Vectorized Latent Diffusion for Generating Driving
                  Simulation Environments},
  booktitle    = {{CVPR}},
  year         = {2025}
}

@inproceedings{SeqGrowGraph,
  author       = {Mengwei Xie and
                  Shuang Zeng and
                  Xinyuan Chang and
                  Xinran Liu and
                  Zheng Pan and
                  Mu Xu and
                  Xing Wei},
  title        = {SeqGrowGraph: Learning Lane Topology as a Chain of Graph Expansions},
  booktitle    = {{ICCV}},
  year         = {2025}
}

@inproceedings{SLEDGE,
  author       = {Kashyap Chitta and
                  Daniel Dauner and
                  Andreas Geiger},
  title        = {{SLEDGE:} Synthesizing Driving Environments with Generative Models
                  and Rule-Based Traffic},
  booktitle    = {{ECCV}},
  year         = {2024}
}

@article{SEPT,
  author       = {Muleilan Pei and
                  Jiayao Shan and
                  Peiliang Li and
                  Jieqi Shi and
                  Jing Huo and
                  Yang Gao and
                  Shaojie Shen},
  title        = {{SEPT:} Standard-Definition Map Enhanced Scene Perception and Topology Reasoning for Autonomous Driving},
  journal      = {IEEE RAL},
  year         = {2025}
}

@inproceedings{TopoLogic,
  author       = {Yanping Fu and
                  Wenbin Liao and
                  Xinyuan Liu and
                  Hang Xu and
                  Yike Ma and
                  Yucheng Zhang and
                  Feng Dai},
  title        = {TopoLogic: An Interpretable Pipeline for Lane Topology Reasoning on
                  Driving Scenes},
  booktitle    = {NeurIPS},
  year         = {2024}
}

@inproceedings{TopoMLP,
  author       = {Dongming Wu and
                  Jiahao Chang and
                  Fan Jia and
                  Yingfei Liu and
                  Tiancai Wang and
                  Jianbing Shen},
  title        = {TopoMLP: {A} Simple yet Strong Pipeline for Driving Topology Reasoning},
  booktitle    = {{ICLR}},
  year         = {2024}
}

@article{TopoNet,
  title={Graph-based topology reasoning for driving scenes},
  author={Li, Tianyu and Chen, Li and Wang, Huijie and Li, Yang and Yang, Jiazhi and Geng, Xiangwei and Jiang, Shengyin and Wang, Yuting and Xu, Hang and Xu, Chunjing and others},
  journal={arXiv preprint arXiv:2304.05277},
  year={2023}
}

@inproceedings{TopoPoint,
  author       = {Yanping Fu and
                  Xinyuan Liu and
                  Tianyu Li and
                  Yike Ma and
                  Yucheng Zhang and
                  Feng Dai},
  title        = {TopoPoint: Enhance Topology Reasoning via Endpoint Detection in Autonomous
                  Driving},
  booktitle    = {NeurIPS},
  year         = {2025}
}

@inproceedings{LaneDiffusion,
  author       = {Zijie Wang and
                  Weiming Zhang and
                  Wei Zhang and
                  Xiao Tan and
                  Hongxing Liu and
                  Yaowei Wang and
                  Guanbin Li},
  title        = {LaneDiffusion: Improving Centerline Graph Learning via Prior Injected
                  {BEV} Feature Generation},
  booktitle    = {{ICCV}},
  year         = {2025}
}

@article{LLamaGen,
  title={Autoregressive Model Beats Diffusion: Llama for Scalable Image Generation},
  author={Sun, Peize and Jiang, Yi and Chen, Shoufa and Zhang, Shilong and Peng, Bingyue and Luo, Ping and Yuan, Zehuan},
  journal={arXiv preprint arXiv:2406.06525},
  year={2024}
}

@article{qwen2,
  title={Qwen2-vl: Enhancing vision-language model's perception of the world at any resolution},
  author={Wang, Peng and Bai, Shuai and Tan, Sinan and Wang, Shijie and Fan, Zhihao and Bai, Jinze and Chen, Keqin and Liu, Xuejing and Wang, Jialin and Ge, Wenbin and others},
  journal={arXiv preprint arXiv:2409.12191},
  year={2024}
}

@inproceedings{resnet,
  author={He, Kaiming and Zhang, Xiangyu and Ren, Shaoqing and Sun, Jian},
  booktitle={{CVPR}}, 
  title={Deep Residual Learning for Image Recognition}, 
  year={2016},
}

@inproceedings{simpleBEV,
  author={Harley, Adam W. and Fang, Zhaoyuan and Li, Jie and Ambrus, Rares and Fragkiadaki, Katerina},
  booktitle={{ICRA}}, 
  title={Simple-BEV: What Really Matters for Multi-Sensor BEV Perception?}, 
  year={2023},
}

@inproceedings{flowtok,
  author    = {Ju He and Qihang Yu and Qihao Liu and Liang-Chieh Chen},
  title     = {FlowTok: Flowing Seamlessly Across Text and Image Tokens},
  booktitle    = {{ICCV}},
  year         = {2025}
}

@inproceedings{CrossFlow,
  author       = {Qihao Liu and
                  Xi Yin and
                  Alan L. Yuille and
                  Andrew Brown and
                  Mannat Singh},
  title        = {Flowing from Words to Pixels: {A} Noise-Free Framework for Cross-Modality
                  Evolution},
  booktitle    = {{CVPR}},
  year         = {2025}
}

@inproceedings{STSU,
  title={Structured Bird's-Eye-View Traffic Scene Understanding from Onboard Images},
  author={Can, Yigit Baran and Liniger, Alexander and Paudel, Danda Pani and Van Gool, Luc},
  booktitle={ICCV},
  year={2021}
}

@inproceedings{CenterLineDet,
  title={CenterLineDet: CenterLine Graph Detection for Road Lanes with Vehicle-mounted Sensors by Transformer for HD Map Generation},
  author={Xu, Zhenhua and Liu, Yuxuan and Sun, Yuxiang and Liu, Ming and Wang, Lujia},
  booktitle={ICRA},
  year={2023}
}

@article{Topo2D,
  title={Enhancing 3D Lane Detection and Topology Reasoning with 2D Lane Priors},
  author={Li, Han and Huang, Zehao and Wang, Zitian and Rong, Wenge and Wang, Naiyan and Liu, Si},
  journal={arXiv preprint arXiv:2406.03105},
  year={2024}
}

@inproceedings{LaneGAP,
  title={Lane Graph as Path: Continuity-preserving Path-wise Modeling for Online Lane Graph Construction},
  author={Liao, Bencheng and Chen, Shaoyu and Jiang, Bo and Cheng, Tianheng and Zhang, Qian and Liu, Wenyu and Huang, Chang and Wang, Xinggang},
  booktitle={ECCV},
  year={2024}
}

@inproceedings{CGNet,
  title={Continuity Preserving Online CenterLine Graph Learning},
  author={Han, Yunhui and Yu, Kun and Li, Zhiwei},
  booktitle={ECCV},
  year={2024}
}

@inproceedings{RoadPainter,
  title={RoadPainter: Points Are Ideal Navigators for Topology TransformER},
  author={Ma, Zhongxing and Liang, Shuang and Wen, Yongkun and Lu, Weixin and Wan, Guowei},
  booktitle={ECCV},
  year={2024}
}

@inproceedings{TSTGT,
  title={Driving Scene Understanding with Traffic Scene-Assisted Topology Graph Transformer},
  author={Rong, Fu and Peng, Wenjin and Lan, Meng and Zhang, Qian and Zhang, Lefei},
  booktitle={ACM MM},
  year={2024}
}

@inproceedings{TopoFormer,
  title={T2SG: Traffic Topology Scene Graph for Topology Reasoning in Autonomous Driving},
  author={Lv, Changsheng and Qi, Mengshi and Liu, Liang and Ma, Huadong},
  booktitle={CVPR},
  year={2025}
}

@inproceedings{RATopo,
  title={RATopo: Improving Lane Topology Reasoning via Redundancy Assignment},
  author={Li, Han and Huang, Shaofei and Xu, Longfei and Gao, Yulu and Mu, Beipeng and Liu, Si},
  booktitle={ACM MM},
  year={2025}
}

@inproceedings{lu2023translating,
  title={Translating Images to Road Network: A Non-Autoregressive Sequence-to-Sequence Approach},
  author={Lu, Jiachen and Peng, Renyuan and Cai, Xinyue and Xu, Hang and Li, Hongyang and Wen, Feng and Zhang, Wei and Zhang, Li},
  booktitle={ICCV},
  year={2023}
}

@article{lu2025translating,
  title={Translating Images to Road Network: A Sequence-to-Sequence Perspective},
  author={Lu, Jiachen and Nie, Ming and Zhang, Bozhou and Peng, Renyuan and Cai, Xinyue and Xu, Hang and Li, Hongyang and Wen, Feng and Zhang, Wei and Zhang, Li},
  journal={IEEE TPAMI},
  year={2025}
}

@inproceedings{LaneGraph2Seq,
  title={LaneGraph2Seq: Lane Topology Extraction with Language Model via Vertex-Edge Encoding and Connectivity Enhancement},
  author={Peng, Renyuan and Cai, Xinyue and Xu, Hang and Lu, Jiachen and Wen, Feng and Zhang, Wei and Zhang, Li},
  booktitle={AAAI},
  year={2024}
}

@inproceedings{Topo2Seq,
  title={Topo2Seq: Enhanced Topology Reasoning via Topology Sequence Learning},
  author={Yang, Yiming and Luo, Yueru and He, Bingkun and Li, Erlong and Cao, Zhipeng and Zheng, Chao and Mei, Shuqi and Li, Zhen},
  booktitle={AAAI},
  year={2025}
}

@inproceedings{Transformer,
  title={Attention is all you need},
  author={Vaswani, Ashish and Shazeer, Noam and Parmar, Niki and Uszkoreit, Jakob and Jones, Llion and Gomez, Aidan N and Kaiser, {\L}ukasz and Polosukhin, Illia},
  booktitle={NeurIPS},
  year={2017}
}

@inproceedings{SMERF,
  title={Augmenting Lane Perception and Topology Understanding with
Standard Definition Navigation Maps},
  author={Luo, Katie Z and Weng, Xinshuo and Wang, Yan and Wu, Shuang and Li, Jie and Weinberger, Kilian Q and Wang, Yue and Pavone, Marco},
  booktitle={ICRA},
  year={2024}
}

@article{TrajTopo,
  title={Enhancing Lane Segment Perception and Topology Reasoning with Crowdsourcing Trajectory Priors},
  author={Jia, Peijin and Luo, Ziang and Wen, Tuopu and Yang, Mengmeng and Jiang, Kun and Cui, Le and Yang, Diange},
  journal={IEEE RAL},
  year={2025}
}

@inproceedings{SMART,
  title={SMART: Advancing Scalable Map Priors for Driving Topology Reasoning},
  author={Ye, Junjie and Paz, David and Zhang, Hengyuan and Guo, Yuliang and Huang, Xinyu and Christensen, Henrik I and Wang, Yue and Ren, Liu},
  booktitle={ICRA},
  year={2025}
}

@inproceedings{Score,
  title={Coherent Online Road Topology Estimation and Reasoning with Standard-Definition Maps},
  author={Pham, Khanh Son and Witte, Christian and Behley, Jens and Betz, Johannes and Stachniss, Cyrill},
  booktitle={IROS},
  year={2025}
}

@article{PriorDrive,
  title={PriorDrive: Enhancing Online HD Mapping with Unified Vector Priors},
  author={Zeng, Shuang and Chang, Xinyuan and Liu, Xinran and Yuan, Yujian and Liang, Shiyi and Pan, Zheng and Xu, Mu and Wei, Xing},
  journal={arXiv preprint arXiv:2409.05352},
  year={2024}
}

@article{MapLite2,
  title={MapLite 2.0: Online HD Map Inference Using a Prior SD Map},
  author={Ort, Teddy and Walls, Jeffrey M and Parkison, Steven A and Gilitschenski, Igor and Rus, Daniela},
  journal={IEEE RAL},
  year={2022}
}

@inproceedings{SatForHDMap,
  title={Complementing Onboard Sensors with Satellite Maps: A New Perspective for HD Map Construction},
  author={Gao, Wenjie and Fu, Jiawei and Shen, Yanqing and Jing, Haodong and Chen, Shitao and Zheng, Nanning},
  booktitle={ICRA},
  year={2024}
}

@article{DiffMap,
  title={DiffMap: Enhancing Map Segmentation With Map Prior Using Diffusion Model},
  author={Jia, Peijin and Wen, Tuopu and Luo, Ziang and Yang, Mengmeng and Jiang, Kun and Liu, Ziyuan and Tang, Xuewei and Lei, Zhiquan and Cui, Le and Zhang, Bo and others},
  journal={IEEE RAL},
  year={2024}
}

@inproceedings{DifFUSER,
  title={Diffusion Model for Robust Multi-sensor Fusion in 3D Object Detection and BEV Segmentation},
  author={Le, Duy-Tho and Shi, Hengcan and Cai, Jianfei and Rezatofighi, Hamid},
  booktitle={ECCV},
  year={2024}
}

@inproceedings{MapDiffusion,
  title={MapDiffusion: Generative Diffusion for Vectorized Online HD Map Construction and Uncertainty Estimation in Autonomous Driving},
  author={Monninger, Thomas and Zhang, Zihan and Mo, Zhipeng and Anwar, Md Zafar and Staab, Steffen and Ding, Sihao},
  booktitle={IROS},
  year={2025}
}

@inproceedings{LaneGNN,
  author       = {Martin B{\"{u}}chner and
                  Jannik Z{\"{u}}rn and
                  Ion{-}George Todoran and
                  Abhinav Valada and
                  Wolfram Burgard},
  title        = {Learning and Aggregating Lane Graphs for Urban Automated Driving},
  booktitle    = {{CVPR}},
  year         = {2023}
}

@inproceedings{Aerial_Image2Map,
  author={He, Songtao and Balakrishnan, Hari},
  title={Lane-Level Street Map Extraction from Aerial Imagery}, 
  booktitle={{WACV}}, 
  year={2022},
}

@inproceedings{waymo,
  author       = {Scott Ettinger and
                  Shuyang Cheng and
                  Benjamin Caine and
                  Chenxi Liu and
                  Hang Zhao and
                  Sabeek Pradhan and
                  Yuning Chai and
                  Ben Sapp and
                  Charles R. Qi and
                  Yin Zhou and
                  Zoey Yang and
                  Aurelien Chouard and
                  Pei Sun and
                  Jiquan Ngiam and
                  Vijay Vasudevan and
                  Alexander McCauley and
                  Jonathon Shlens and
                  Dragomir Anguelov},
  title        = {Large Scale Interactive Motion Forecasting for Autonomous Driving
                  : The Waymo Open Motion Dataset},
  booktitle    = {{ICCV}},
  year         = {2021}
}

@article{av2,
  title={Argoverse 2: Next generation datasets for self-driving perception and forecasting},
  author={Wilson, Benjamin and Qi, William and Agarwal, Tanmay and Lambert, John and Singh, Jagjeet and Khandelwal, Siddhesh and Pan, Bowen and Kumar, Ratnesh and Hartnett, Andrew and Pontes, Jhony Kaesemodel and others},
  journal={arXiv preprint arXiv:2301.00493},
  year={2023}
}

@article{nuplan,
  author       = {Holger Caesar and
                  Juraj Kabzan and
                  Kok Seang Tan and
                  Whye Kit Fong and
                  Eric M. Wolff and
                  Alex H. Lang and
                  Luke Fletcher and
                  Oscar Beijbom and
                  Sammy Omari},
  title        = {nuPlan: {A} closed-loop ML-based planning benchmark for autonomous
                  vehicles},
  journal={arXiv preprint arXiv:2106.11810},
  year={2021}
}

@article{LaneGraphNet,
  author       = {Jannik Z{\"{u}}rn and
                  Johan Vertens and
                  Wolfram Burgard},
  title        = {Lane Graph Estimation for Scene Understanding in Urban Driving},
  journal      = {IEEE RAL},
  year         = {2021}
}

@inproceedings{DiT,
  author       = {William Peebles and
                  Saining Xie},
  title        = {Scalable Diffusion Models with Transformers},
  booktitle    = {{ICCV}},
  year         = {2023}
}

@inproceedings{vectormapnet,
        title={VectorMapNet: End-to-end Vectorized HD Map Learning},
        author={Liu, Yicheng and Yuan, Tianyuan and Wang, Yue and Wang, Yilun and Zhao, Hang},
        booktitle={{ICML}},
        year={2023},
}

@article{li2026unified,
  title={Unified Modeling of Lane and Lane Topology for Driving Scene Reasoning},
  author={Li, Han and Gao, Yulu and Liu, Si and Wang, Yuhang and Liu, Bo and Mu, Beipeng},
  journal={IEEE TCSVT},
  year={2026},
  publisher={IEEE}
}

@inproceedings{MV2D,
  title={Object as query: Lifting any 2d object detector to 3d detection},
  author={Wang, Zitian and Huang, Zehao and Fu, Jiahui and Wang, Naiyan and Liu, Si},
  booktitle={ICCV},
  year={2023}
}

@inproceedings{ECFusion,
  title={Eliminating Cross-modal Conflicts in BEV Space for LiDAR-Camera 3D Object Detection},
  author={Fu, Jiahui and Gao, Chen and Wang, Zitian and Yang, Lirong and Wang, Xiaofei and Mu, Beipeng and Liu, Si},
  booktitle={ICRA},
  year={2024}
}

@inproceedings{CoGMP,
  title={Generative map priors for collaborative BEV semantic segmentation},
  author={Fu, Jiahui and Gong, Yue and Wang, Luting and Zhang, Shifeng and Zhou, Xu and Liu, Si},
  booktitle={CVPR},
  year={2025}
}

@inproceedings{wang2024image,
  title={Image understanding makes for a good tokenizer for image generation},
  author={Wang, Luting and Zhao, Yang and Zhang, Zijian and Feng, Jiashi and Liu, Si and Kang, Bingyi},
  booktitle={NeurIPS},
  year={2024}
}
\end{document}